\documentclass[journal]{IEEEtran}

\ifCLASSINFOpdf
\else
   \usepackage[dvips]{graphicx}
\fi

\usepackage{url}
\usepackage{graphicx}
\usepackage[caption=false]{subfig}
\newcommand{\squeezeup}{\vspace{-1.6mm}}
\usepackage{booktabs}
\usepackage{algorithm}
\usepackage[table,xcdraw]{xcolor}
\usepackage[hang,flushmargin]{footmisc} 
\usepackage{booktabs}
\usepackage{amsfonts}
\usepackage{amssymb}
\usepackage{amsmath}
\usepackage{bm,bbm}
\usepackage{hyperref}
\usepackage[noend]{algpseudocode}
\newcommand\Algphase[1]{%
\vspace*{-0.5\baselineskip}\Statex\hspace*{\dimexpr-\algorithmicindent-2pt\relax}\rule{8.56cm}{0.4pt}%
\Statex\hspace*{-\algorithmicindent}\textbf{#1}%
\vspace*{-0.5\baselineskip}\Statex\hspace*{\dimexpr-\algorithmicindent-2pt\relax}\rule{8.56cm}{0.4pt}%
}
\definecolor{dblue}{RGB}{0,0,0}
\definecolor{blue}{RGB}{0,0,255}
\newcommand{\rev}[1]{{\color{dblue} #1}}

\begin{document}

\title{\rev{UAVM: Towards Unifying Audio and Visual Models}}

\author{Yuan Gong,~\IEEEmembership{Member,~IEEE,}
        Alexander H. Liu,
        Andrew Rouditchenko,
        and James Glass,~\IEEEmembership{Fellow, IEEE}
\thanks{Code at \href{https://github.com/yuangongnd/uavm}{\texttt{\color{blue}{github.com/yuangongnd/uavm}}}. The authors are with the Computer Science and Artificial Intelligence Laboratory, Massachusetts Institute of Technology, USA (e-mail: \{yuangong, glass\}@mit.edu).}}

\markboth{IEEE Signal Processing Letters}
{Shell \MakeLowercase{\textit{et al.}}: Bare Demo of IEEEtran.cls for IEEE Journals}
\maketitle

\begin{abstract}

\rev{Conventional audio-visual models have independent audio and video branches. In this work, we \emph{unify} the audio and visual branches by designing a \underline{U}nified \underline{A}udio-\underline{V}isual \underline{M}odel (UAVM). The UAVM achieves a new state-of-the-art audio-visual event classification accuracy of 65.8\% on VGGSound. More interestingly, we also find a few intriguing properties of UAVM that the modality-independent counterparts do not have.}

\end{abstract}

\begin{IEEEkeywords}
audio-visual learning, unified model
\end{IEEEkeywords}

\IEEEpeerreviewmaketitle
\section{Introduction}


Humans perceive and understand their world by combining different sensory input modalities including sound and vision. To enable AI systems to have a similar ability, \emph{audio-visual multi-modal learning} has been extensively studied~\cite{ramachandram2017deep,zhu2021deep}.
Due to the inherent differences between audio and video, conventional audio-visual learning methods typically either use handcrafted modality-specific features or have two \emph{independent} audio and visual branches with different \emph{architectures}, \emph{training schemes}, and \emph{model weights}~\cite{aytar2016soundnet,arandjelovic2017look,rouditchenko2021avlnet,monfort2021spoken,chang2022on}.

\rev{This modality-specific paradigm began to change after the Transformer~\cite{vaswani2017attention} showed its effectiveness for various tasks and modalities including audio~\cite{gong21b_interspeech,gong2022ssast,ao2022speecht5} and video~\cite{dosovitskiy2020image,touvron2021training}.}
Specifically, the Perceiver~\cite{jaegle2021perceiver} model shows that it is possible to handle arbitrary configurations of different modalities using a \emph{unified model architecture}, although the training method is still modality-specific. In addition to unified model architecture, data2vec~\cite{baevski2022data2vec} further demonstrates that a \emph{unified training scheme} can be applied to different modalities to obtain state-of-the-art performance. Despite the unified model architecture and training scheme, the models of different modalities are trained independently and have independent weights. To go one step even further, SkillNet~\cite{dai2022one}, EAO~\cite{shvetsova2022everything}, VATT~\cite{akbari2021vatt}, and PolyViT~\cite{likhosherstov2021polyvit} share part of the \emph{model weights} among different modalities. 
Specifically, SkillNet~\cite{dai2022one} and EAO~\cite{shvetsova2022everything} are single models that handle multiple modalities, but different parts of the model weights are specialized for processing different modalities. 
VATT~\cite{akbari2021vatt} uses a modality-agnostic, single-backbone Transformer and shares weights among different modalities. However, the training scheme of VATT is modality-asymmetric. PolyViT~\cite{likhosherstov2021polyvit} shares model weights except the input tokenizer and task head.

\begin{figure}[t]
\centering
\includegraphics[width=8.cm]{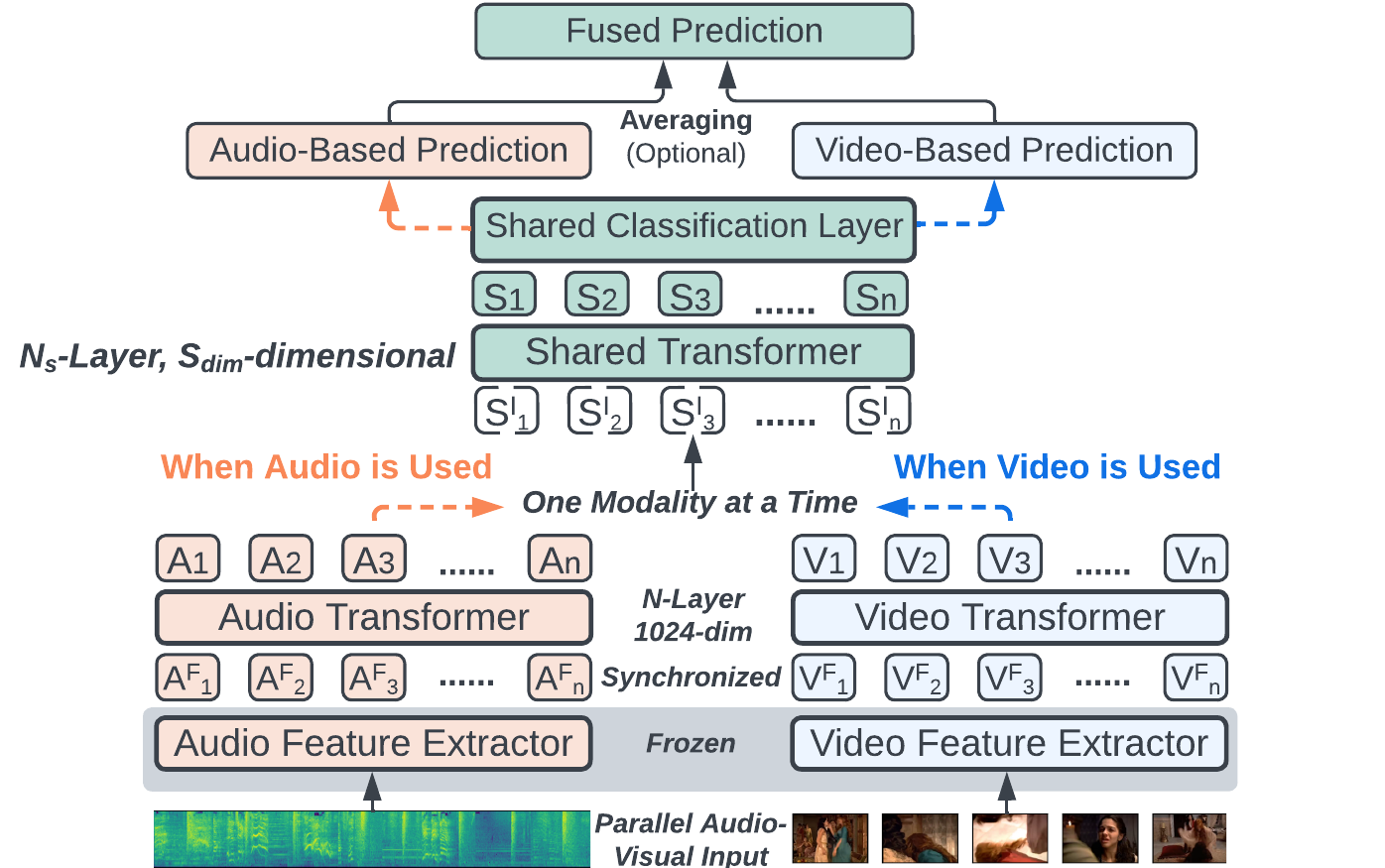}
\squeezeup\squeezeup
\caption{An illustration of the unified audio-visual model (UAVM).}
\label{fig:illustration}
\squeezeup\squeezeup\squeezeup
\end{figure}


The motivations for cross-modality model unification are multifold. First, it minimizes the use of handcrafted priors and inductive biases for each individual modality, which is more data-driven and saves manual effort. Second, a unified model that handles multiple modalities can be more parameter-efficient than a set of modality-specific models. Third, unified models are ideal \emph{foundation models}~\cite{Bommasani2021FoundationModels} that can be adapted to a wide range of downstream tasks of multiple modalities.

\rev{However, how unified models differ from modal-specific models remains unclear. Are unified models just two modal-independent models ``glued'' together? Does a unified model indeed encode audio and visual input into a unified representation space? We answer these questions by building and analyzing a \underline{U}nified \underline{A}udio-\underline{V}isual \underline{M}odel (UAVM) that unifies 1) model architecture; 2) weights for high layers including the decision layer; and 3) training with a unified algorithm for audio and video. On VGGSound~\cite{chen2020vggsound}, UAVM achieves a new state-of-the-art accuracy for audio-visual event classification.}

\section{The Unified Audio-Visual Model}
\subsection{UAVM Model Architecture}
\label{sec:mdl_arc}

The unique aspect of the UAVM is the shared Transformer and classification layer whose weights are shared between different modalities.  This feature enables the UAVM to process both audio \emph{and} video independently. Before the shared Transformer, we still have modal-specific feature extractors and optional modal-specific Transformers for each modality.

As shown in Figure~\ref{fig:illustration}, the audio or video is first input to the corresponding modality-specific feature extractor. Both audio and video feature extractors are ConvNeXt-Base~\cite{liu2022convnet}. For video, we use ImageNet pretrained ConvNeXt as the feature extractor. We uniformly sample RGB frames from the video at 3 FPS, input each frame to the ConvNeXt-Base, and get the mean-pooled penultimate layer representation of 1024-dimensions, i.e., for each 10-second video, the video features are 30 1024-dimensional vectors $\{V^{F}_{1}, ..., V^{F}_{30}\}$. For audio, we train a ConvNeXt using in-domain audio data (AudioSet or VGGSound) with ImageNet initialization~\cite{gong_psla} and then use it as the audio feature extractor. Each 10-second waveform is first converted to a $1000\times128$ log Mel filterbank (fbank) feature vector computed with a 25ms Hanning window every 10ms and input to the feature extractor. The output of the penultimate layer of ConvNeXt is a 30 (time)$\times$4 (frequency)$\times$1024 tensor. We apply a frequency mean pooling to produce 30 1024-dimension audio features $\{A^{F}_{1}, ..., A^{F}_{30}\}$. Note that we intend to make the audio and video features synchronized. Both feature extractors are frozen during training. 

We then input the L2-normalized $\{A^F\}$ or $\{V^F\}$ to corresponding $N$-layer modality-specific Transformers to produce $\{A_1, ..., A_{30}\}$ or $\{V_1, ..., V_{30}\}$. After that, either $\{A_1, ..., A_{30}\}$ or $\{V_1, ..., V_{30}\}$ is used as an input $\{S^I_1, ..., S^I_{30}\}$ to the shared, modality-agnostic Transformer of $N_s$ layers. This approach is fundamentally different from concatenating audio and visual tokens as input to the shared Transformer as in MBT~\cite{nagrani2021attention} and Merlot Reserve~\cite{zellers2022merlot}. We mean-pool the output of the shared Transformer $\{S_1, ..., S_{30}\}$ and input it to a shared linear classification layer. When only one modality is input, we use the output of the shared linear classification layer as the single-modality prediction; when both modalities are input, we run one forward pass for each modality and average the outputs of the two passes as a fused prediction (Algorithm~\ref{alg:train}). In other words, UAVM is robust to missing modalities.

Throughout the experiments, all Transformer layers have 4 attention heads and the modality-specific Transformer always has an embedding dimension of 1024. We tune the embedding dimension of the shared Transformer $S_{dim}$ from 16 to 1024 to control its capacity. We fix the total number of Transformer layers to 6 ($N + N_s=6$), and tune the number of modal-specific Transformer layers, $N$, and shared Transformer layers, $N_s$, from 0 to 6 to control the model unification level. When $N=0$ and $N_s=6$, the audio and visual models are maximally unified; when $N=6$ and $N_s=0$, the audio and visual models are completely independent including the classification layer. We set the model with $N=N_s=3$ and $S_{dim}=1024$ as the base UAVM model. 

\begin{figure}[t]
\vspace{-1em}
\begin{algorithm}[H]
\footnotesize
\caption{UAVM Model Training and Inference}
\label{alg:train}
\begin{algorithmic}[1]
\Require{Dataset $\mathcal{D}=\{A,V,Label\}$, UAVM Model $\mathcal{M}=\{\theta_a,\theta_v,\theta_s\}$}
\Algphase{Training ($\mathcal{D},\mathcal{M},\lambda_{\text{MT}}$)}
\While {$i < max\;training\;iteration$}
\newline $\triangleright$ One modality is used to train the model at an iteration
\If{$unif(0,1)<$ modality training weight $\lambda_{\text{MT}}$}
\State sample a batch of audio $\{A_i, Label_i\}$
\State $Pred_{a}=\mathcal{M}_{\{\theta_a,\theta_s\}}(A_i)$
\State $\mathcal{L} = Loss(Label,Pred_{a})$
\State backpropagate and update $\{\theta_a, \theta_s\}$
\Else
\State sample a batch of video $\{V_i, Label_i\}$
\State $Pred_{v}=\mathcal{M}_{\{\theta_v,\theta_s\}}(V_i)$
\State $\mathcal{L} = Loss(Label,Pred_{v})$
\State backpropagate and update $\{\theta_v, \theta_s\}$
\EndIf
\EndWhile
\Return $\mathcal{M}$

\Algphase{Inference ($\{A, V\},\mathcal{M}$)}
\If{$A\;!=None$ and $V\;!=None$}
\State $Pred=(\mathcal{M}_{\{\theta_a,\theta_s\}}(A)+\mathcal{M}_{\{\theta_v,\theta_s\}}(V))/2$
\ElsIf{one modality is missing}
\State $Pred=\mathcal{M}_{\{\theta_v,\theta_s\}}(V)$ when $A==None$
\State $Pred=\mathcal{M}_{\{\theta_a,\theta_s\}}(A)$ when $V==None$
\EndIf
\Return $Pred$
\end{algorithmic}
\end{algorithm}
\vspace{-2.2em}
\end{figure}

\subsection{UAVM Model Training}
We train UAVM with Algorithm~\ref{alg:train}. In each training iteration, only one modality is used. This is implemented by using a modality training weight $\lambda_{\text{MT}}$ and uniform sampling, i.e., audio and video have $\lambda_{\text{MT}}$ and $1-\lambda_{\text{MT}}$ probability to be used, respectively. In other words, UAVM does not explicitly leverage the audio-visual correspondence information and can be trained with unpaired audio and video data (though in this paper, we focus on parallel datasets for simplicity).  By default, we use modality training weight $\lambda_{\text{MT}}=0.5$. As in prior work~\cite{gong21b_interspeech,gong_psla,nagrani2021attention}, we train the model with mixup~\cite{zhang2018mixup}, balanced sampling, label smoothing, and random time shifts. These training techniques are applied to both modalities with exactly the same hyperparameters, i.e., we use a unified training pipeline for the two modalities.

\squeezeup
\section{Experiment and Discussion}

\subsection{Experiment Settings}

\subsubsection{Dataset}

We use two widely-used datasets for audio and video event classification: AudioSet~\cite{gemmeke2017audio} and VGGSound~\cite{chen2020vggsound}. AudioSet is a collection of 2M 10-second YouTube video clips labeled with the sounds that the clip contains from a set of 527 labels. We downloaded 1.8M training and 17K evaluation audio and video samples for our experiment. VGGSound~\cite{chen2020vggsound} is a collection of 200K 10-second YouTube video clips annotated with 309 classes. We download 184K training and 15K test samples for our experiments. One advantage of VGGSound is that the sound source is always visually present in the video clip. Also, its moderate size allows us to conduct extensive experiments with our computational resources. Therefore, while we compare UAVM performance with existing methods on both AudioSet and VGGSound, we conduct all ablation studies and analyses on VGGSound only.

\subsubsection{Training Details}

For all experiments, we train the model with a batch size of 144 and the Adam optimizer~\cite{kingma2015adam}. For the main experiments, we use an initial learning rate of 1e-5 and 5e-5 for AudioSet and VGGSound, respectively, and decrease the learning rate with a factor of 0.5 every epoch. We train the model for 10 epochs and report the last epoch result. For ablation studies, we tune hyper-parameters to ensure a fair comparison. We repeat all experiments 3 times with different random seeds and report the mean and standard deviation. The standard deviation is shown as shaded areas in Figure~\ref{fig:exp1}-\ref{fig:exp4}.

\squeezeup\squeezeup
\subsection{Model Performance Comparison}

We compare the performance of three base models:

\textbf{1. UAVM}. The base UAVM described in Section~\ref{sec:mdl_arc} with 3 modal-specific Transformer layers and 3 shared Transformer layers ($N=N_s=3$) and $S_{dim}=1024$.

\textbf{2. Modal-Independent Model}. The base UAVM without shared Transformer layers (i.e., $N=6, N_s=0$), so the audio and visual models are completely independent including the classification layer. 

\textbf{3. Cross-Modal Attention Model}. The base UAVM model with 3 modal-specific Transformer layers and 3 shared Transformer layers ($N=N_s=3$) but instead of inputting one modality at a time to the shared Transformer, this model \emph{concatenates} the outputs of modal-specific Transformers $\{A\}$ and $\{V\}$ together and inputs $\{A, V\}$ to the shared Transformer. This model only works when both modalities are input.

We show the results in Table~\ref{tab:main_res}. Key conclusions are as follows: First, compared with the modal-independent model, UAVM achieves almost the same fusion performance when both modalities are input, and even slightly better results when a single modality is input \rev{with 76\% of the parameters}, demonstrating the feasibility of using a single network for two different modalities. Second, comparing UAVM with the ``MBT-style" cross-modal attention model, we find a performance discrepancy between the datasets, i.e., UAVM is noticeably better on VGGSound while the cross-modal attention model is noticeably better on AudioSet. This is potentially because the sound source is always visually present in VGGSound videos but not in AudioSet videos. The UAVM strategy of giving equal weight to both modalities performs better than cross-modal attention models that could be dominated by one modality on tasks like VGGSound (video is always informative), but worse on AudioSet (video is not always informative). Finally, Transformer models with pretrained frozen features are a strong baseline with low computational cost. As a consequence, UAVM achieves a new SOTA performance on VGGSound, and outperforms all previous methods except MBT~\cite{nagrani2021attention} on AudioSet. \rev{Note this is a fair comparison as both UAVM and MBT use ImageNet pretraining}. \rev{Compared with MBT raw patches, the frozen feature input sequence length} is much shorter (30 vs 1,500+), making it more computationally efficient since the Transformer has quadratic complexity.

\begin{table}[t]
\scriptsize
\centering
\caption{Model Performance Comparison on VGGSound and AudioSet. ($^\ast$ single-modal model trained independently. \\$^\dagger$ modality-missing results of a multi-modal model. )}
\vspace{-1.0em}
\label{tab:main_res}
\begin{tabular}{@{}lccc@{}}
\toprule
\multicolumn{1}{c}{\textbf{VGGSound (Top-1 Accuracy, \%)}}        & Audio & Video & Fusion \\ \midrule
Chen \emph{et al.}~\cite{chen2020vggsound}& 48.8                      & -                         & -         \\
AudioSlowFast~\cite{kazakos2021slow}& 50.1                      & -                         & -               \\
MBT~\cite{nagrani2021attention}     & 52.3$^\ast$   & 51.2$^\ast$             & 64.1                       \\ \midrule
Our Cross-Modal Attention Model & -                         & -                         & 62.9$\pm$0.2         \\
Our Modal-Independent Model & 56.5$\pm$0.1$^\ast$              & 49.7$\pm$0.2$^\ast$              & 65.7$\pm$0.2             \\
Our UAVM Model         & \textbf{56.5$\pm$0.1}$^\dagger$   & \textbf{49.9$\pm$0.2}$^\dagger$              & \textbf{65.8$\pm$0.1}    \\ \midrule\midrule
\multicolumn{1}{c}{\textbf{Full AudioSet (mAP)}} & Audio & Video & Fusion \\ \midrule
GBlend~\cite{wang2020makes} & 32.4$^\ast$                     & 18.8$^\ast$                     & 41.8                     \\
Attn Audio-Visual~\cite{fayek2021large} & 38.4$^\ast$         & 25.7$^\ast$                     & 46.2                     \\
Perceiver~\cite{jaegle2021perceiver} & 38.4$^\ast$                     & 25.8$^\ast$                     & 44.2            \\
MBT~\cite{nagrani2021attention} (w/ 500k training samples) & 44.3$^\ast$                     & 32.3$^\ast$                      & 52.1                       \\ \midrule
Our Cross-Modal Attention Model & -                     & -                         & \textbf{50.4$\pm$0.1} \\
Our Modal-Independent Model     & 45.5$\pm$0.0$^\ast$          & 26.8$\pm$0.1$^\ast$              & 48.1$\pm$0.1        \\
Our UAVM Model                    & \textbf{45.6$\pm$0.0}$^\dagger$ & \textbf{27.4$\pm$0.1}$^\dagger$     & 48.0$\pm$0.0        \\ \bottomrule
\end{tabular}
\vspace{-2.5em}
\end{table}

\begin{figure}[t]
\subfloat{\includegraphics[width=8.5cm]{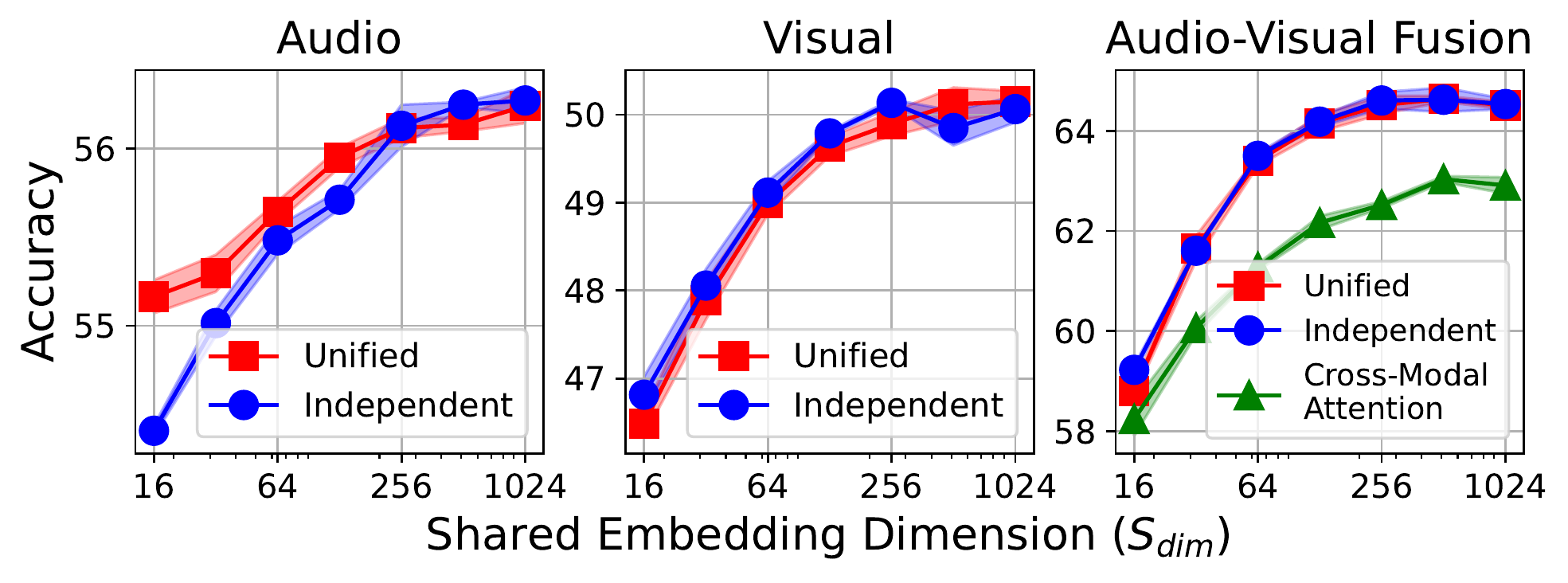}}
\squeezeup\squeezeup

\subfloat{\includegraphics[width=8.5cm]{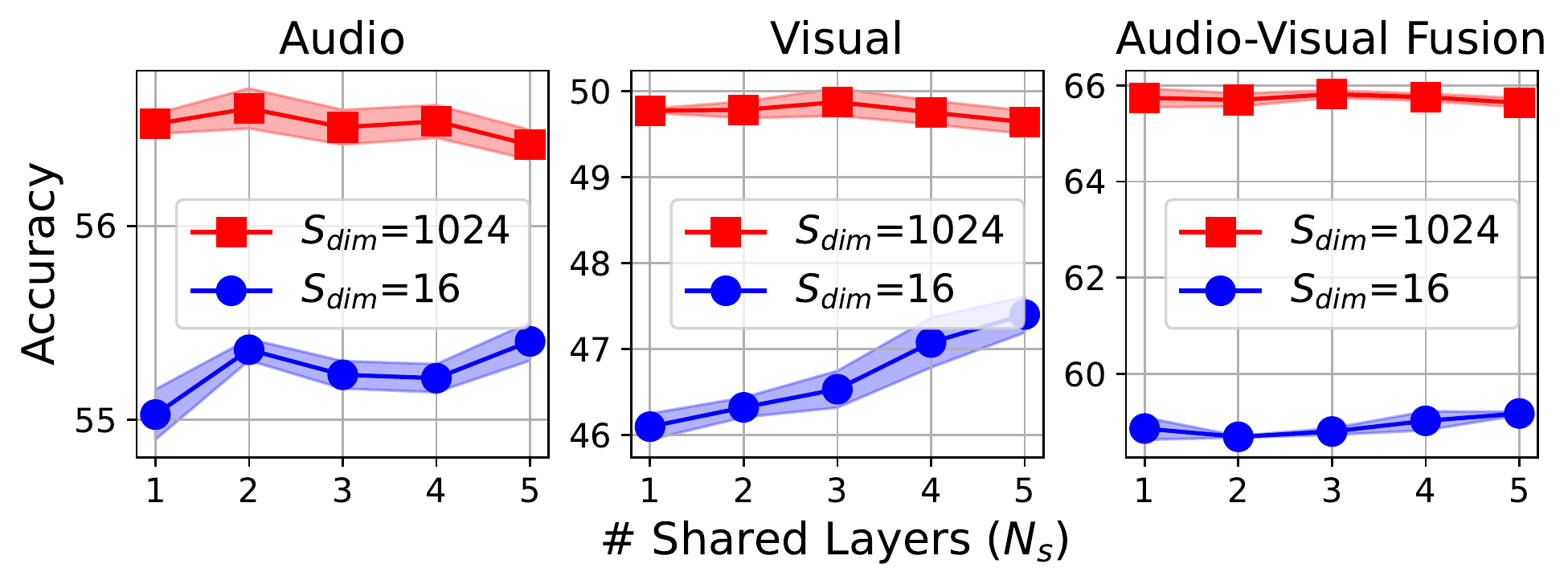}}
\squeezeup\squeezeup

\subfloat{\includegraphics[width=8.5cm]{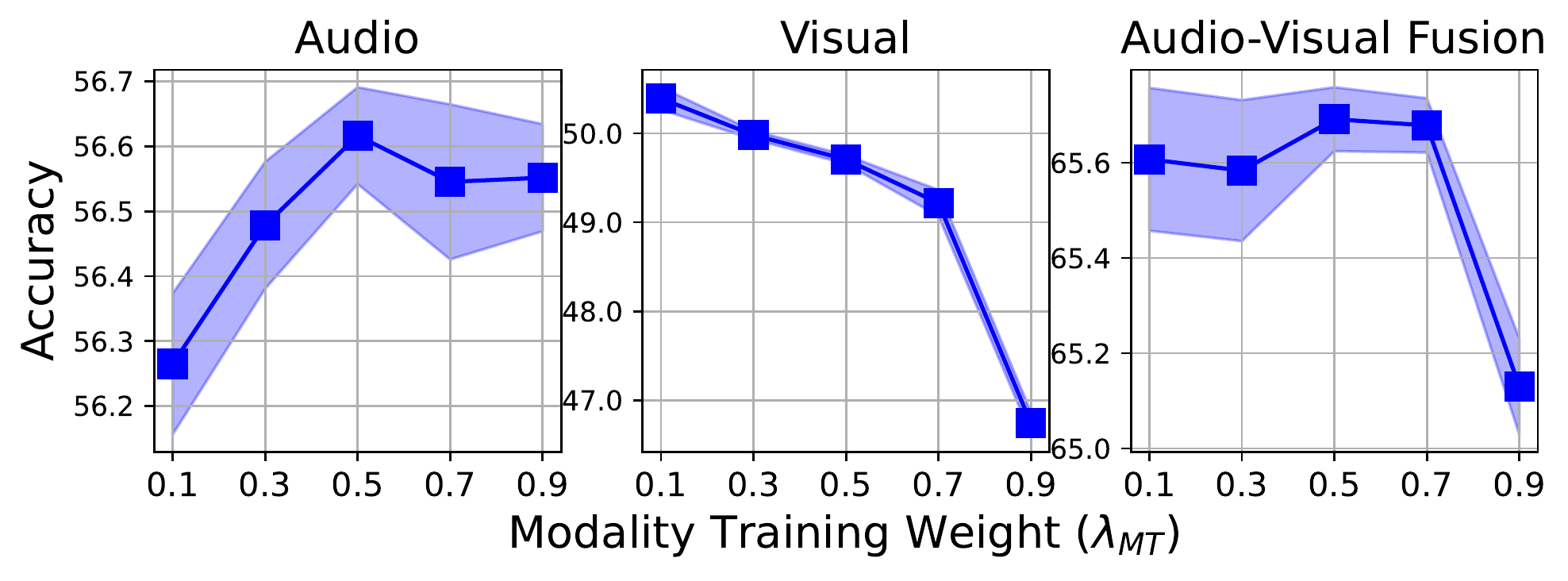}}
\vspace{-1.0em}
\caption{The audio-based, video-based, and fused accuracy on VGGSound with various shared embedding dimension $S_{dim}$ (upper), number of shared layers $N_s$ (middle), and modality training weight $\lambda_{\text{MT}}$ (lower).}
\label{fig:exp1}
\vspace{-1.5em}
\end{figure}

We then conduct a series of ablation studies. We set models with $N=N_s=3$ and $S_{dim}=1024$ as the base UAVM and change one factor at a time to observe the performance change. First, we constrain the shared Transformer embedding dimension $S_{dim}$ to smaller values to lower its capacity for UAVM and compare it with other models. For a fair comparison, the modal-independent and cross-modal attention models also have three 1024-dimensional Transformer layers and three $S_{dim}$-dimensional Transformer layers. With a limited capacity, the shared Transformer is forced to share neurons for two modalities. As shown in Figure~\ref{fig:exp1} (upper), we find model performance generally improves with larger $S_{dim}$, but even when $S_{dim}$ is very small, UAVM still performs similarly or even better than the modal-independent model. Second, we tune the number of shared layers $N_{s}$ to change the level of model unification. Note that we fix the total number of Transformer layers to be 6, i.e., $N + N_s=6$. As shown in Figure~\ref{fig:exp1} (middle), when $S_{dim}$ is small, the single-modality accuracy improves with more shared layers while when $S_{dim}$ is large, the number of shared layers does not impact the performance much. \rev{This is true even when all 6 layers are shared (65.6\% accuracy)}. Finally, as shown in Figure~\ref{fig:exp1} (lower), even though audio and video are not equally informative for the event classification task (reflected in different accuracies), using a 0.5 $\lambda_{\text{MT}}$ leads to optimal fusion results.

\squeezeup\squeezeup
\subsection{Unified Audio-Visual Representation Space}
\label{sec:exp2}

\begin{figure}[t]
\subfloat{\includegraphics[width=9.cm]{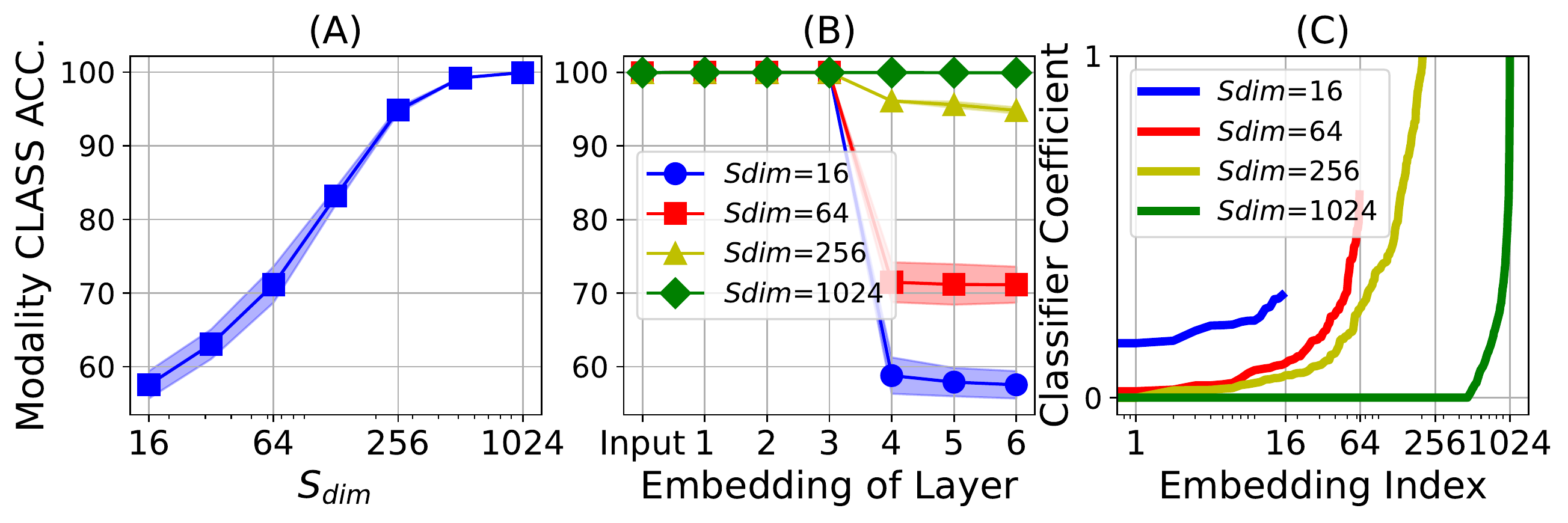}}
\squeezeup\squeezeup

\subfloat{\includegraphics[width=9.cm]{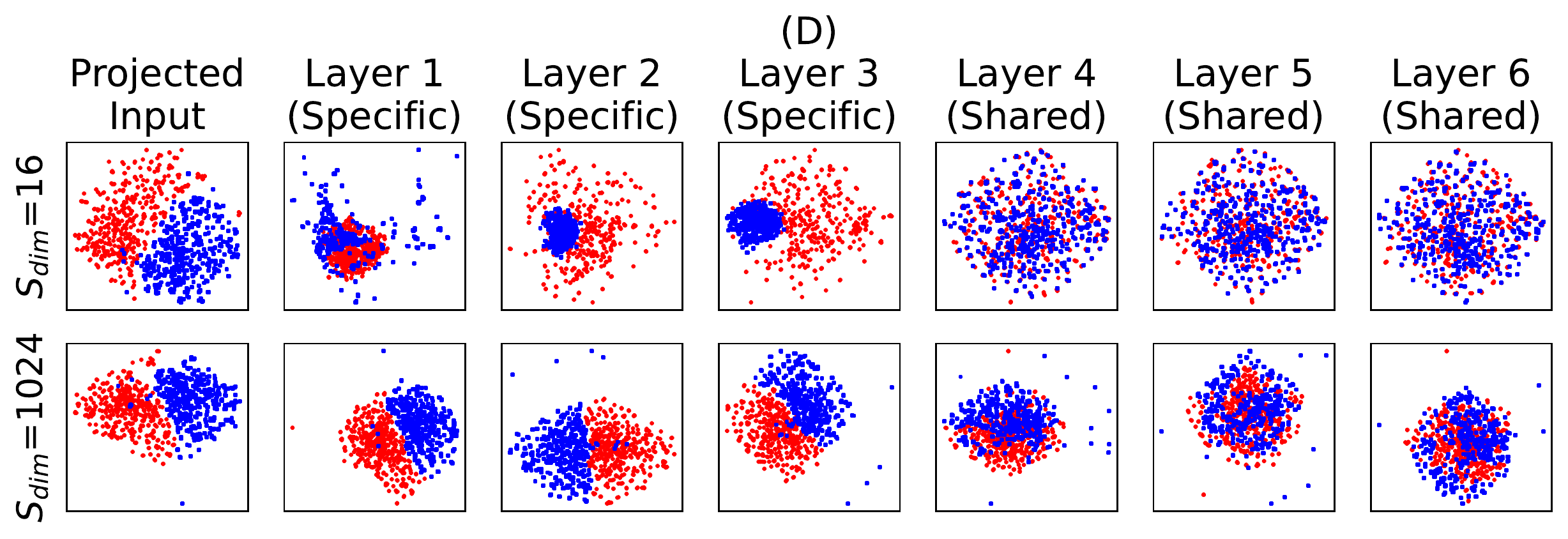}}
\vspace{-1.0em}
\caption{(A) Modality classification accuracy based on the last layer representation of UAVM with various $S_{dim}$. (B) Modality classification accuracy based on intermediate representations of each layer of UAVM models. (C) The sorted modality classifier coefficients. (D) t-SNE plot of the intermediate representations of each layer with audio input (red) and video input (blue). }
\label{fig:exp2}
\vspace{-1.0em}
\end{figure}

One core question about the unified model is if it indeed encodes two modalities in a unified latent space, or just processes each modality with a part of its parameters. We explore this by checking if a logistic regression model can successfully classify the input modality based on the mean-pooled penultimate layer representation of the UAVM model. Specifically, we use half of the VGGSound test set to train a logistic regression model and use the other half for testing. As shown in Figure~\ref{fig:exp2}.A, when the shared Transformer has a small $S_{dim}$ and limited capacity, a logistic regression model is unable to predict the input modality based on the penultimate layer representation, in other words, the two modalities are encoded in a unified space. However, the modality classification accuracy gradually increases with $S_{dim}$, indicating that the model, though with shared weights, tends to encode two modalities in separate spaces when it has redundant capacity. Nevertheless, we do not see one or a small number of dimensions of the representation specifically encoding the input modality information because the classifier does not have a dominant coefficient (Figure~\ref{fig:exp2}.C). Further, as shown in Figure~\ref{fig:exp2}.B, the modality of the input and modal-specific layer (layer 1-3) representations can be perfectly classified but the modality classification accuracy drops suddenly with the first shared layer (layer 4) representation, indicating that it is the shared Transformer that maps the two very different inputs to a unified space. The above findings can also be confirmed with the t-SNE plot of the representations of each layer with audio and video input shown in Figure~\ref{fig:exp2}.D.

\begin{figure}[]
\centering
\includegraphics[width=8.cm]{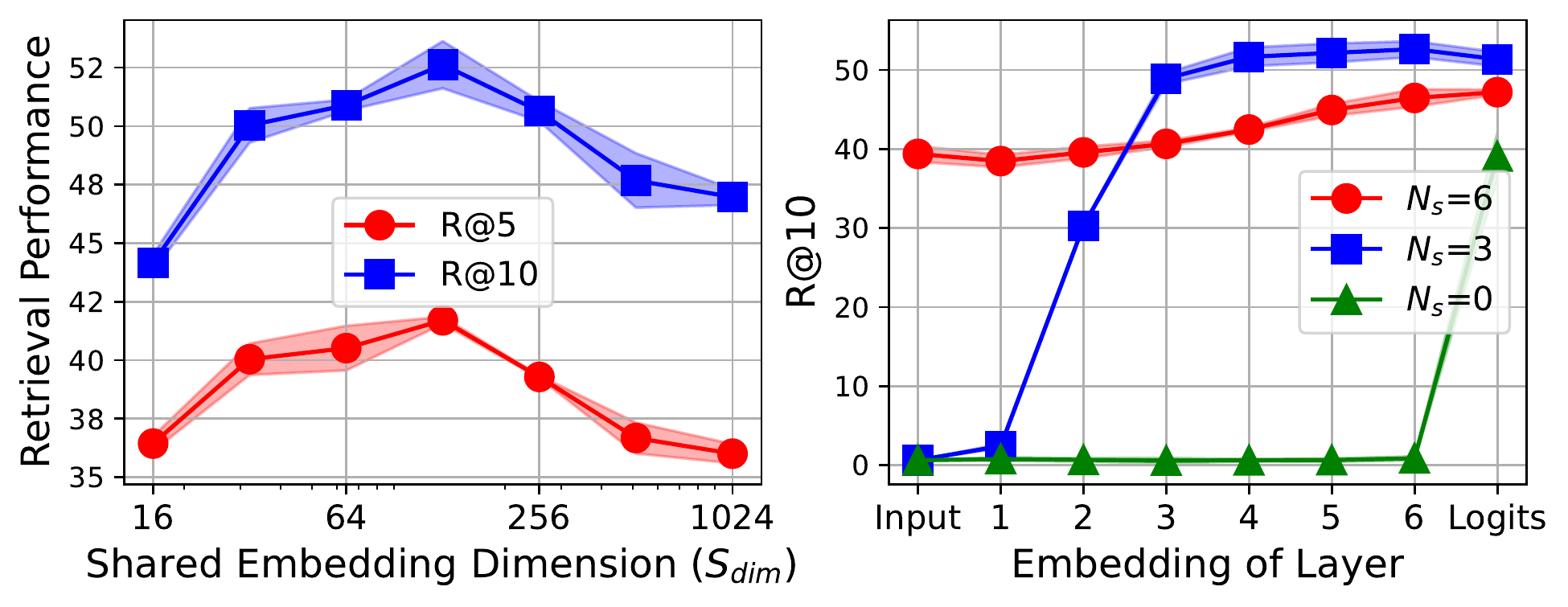}
\vspace{-1.0em}
\caption{(Left) A-V retrieval performance based on the last shared Transformer layer representation of UAVM models ($N_s=3$) with various shared embedding dimensions $S_{dim}$. (Right) A-V retrieval performance based on the representation of various layers of models with $N_s=0$ (fully independent model), $N_s=3$, and $N_s=6$ (fully unified model) and $S_{dim}=128$.}
\label{fig:exp3}
\vspace{-1.2em}
\end{figure}

\squeezeup\squeezeup
\subsection{Audio-Visual Representation Correspondence}

\rev{One further question is how UAVM \emph{aligns} paired audio and visual input in the latent representation space. Interestingly, we find a clear discrepancy between UAVM and modal-independent models.}
As a probing task, we calculate the A-V retrieval recall based on cosine similarity on a 1.5k subset of the VGGSound evaluation set (309 classes, 5 samples per class). 
Note that with reasonable single-modal classification accuracies, the audio and video of the same class can be naturally retrieved from each other. However, as shown in Figure~\ref{fig:exp3} (right), for the modal-independent model ($N_s=0$), the A-V retrieval recall is close to 0 for representations of all layers except the final linear classification head (i.e., the prediction logits) while for UAVMs, the A-V retrieval recall is much higher for representations for front layers (the more shared layers, the earlier a high A-V retrieval recall is achieved), indicating that the shared Transformer layers \rev{tend to} propagate the supervision signal to front layers. \rev{Nonetheless, the unified models learn A-V correspondences beyond just intra-class correspondence as the A-V retrieval recall of UAVMs ($S_{dim}=128$) is about 20\% higher than the modal-independent model, while their classification accuracies are similar. This is particularly interesting because UAVMs do not see simultaneous audio and video pairs, nor has any explicit audio-visual correspondence loss been applied during training. 
Further, as shown in Figure~\ref{fig:exp3} (left), when $S_{dim}>128$, the A-V retrieval recall starts to decrease with the increase of shared embedding dimension $S_{dim}$, while the classification accuracy consistently improves with larger $S_{dim}$ (Figure~\ref{fig:exp1}, upper), indicating UAVM with redundant capacity is worse in audio-visual alignment and closer to modal-independent models, which is consistent with the finding in Section~\ref{sec:exp2}.
}

\begin{figure}[t]
\minipage{2.4cm}
  \includegraphics[width=2.5cm]{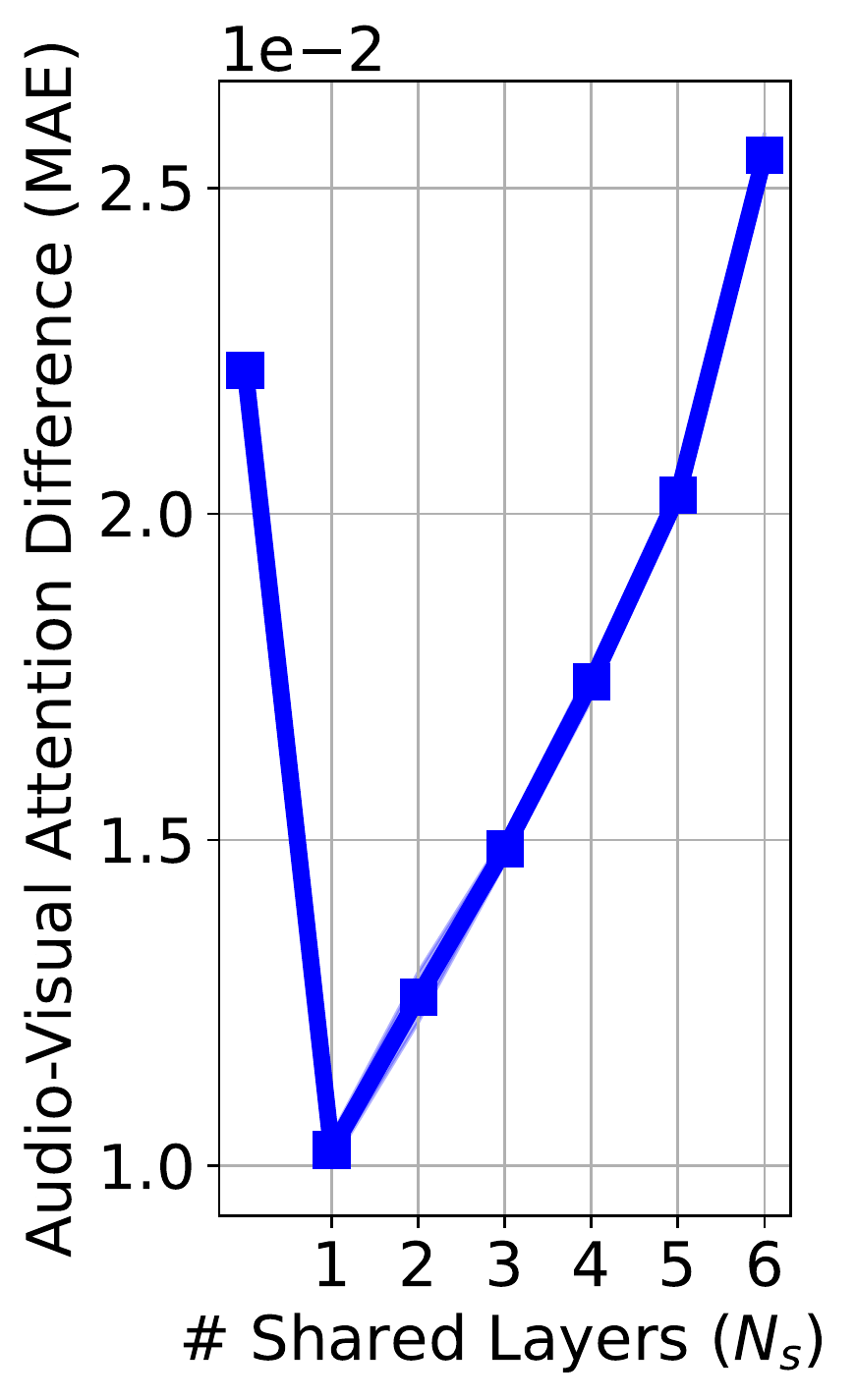}
\endminipage\hfill
\minipage{6.5cm}
  \includegraphics[width=6.5cm]{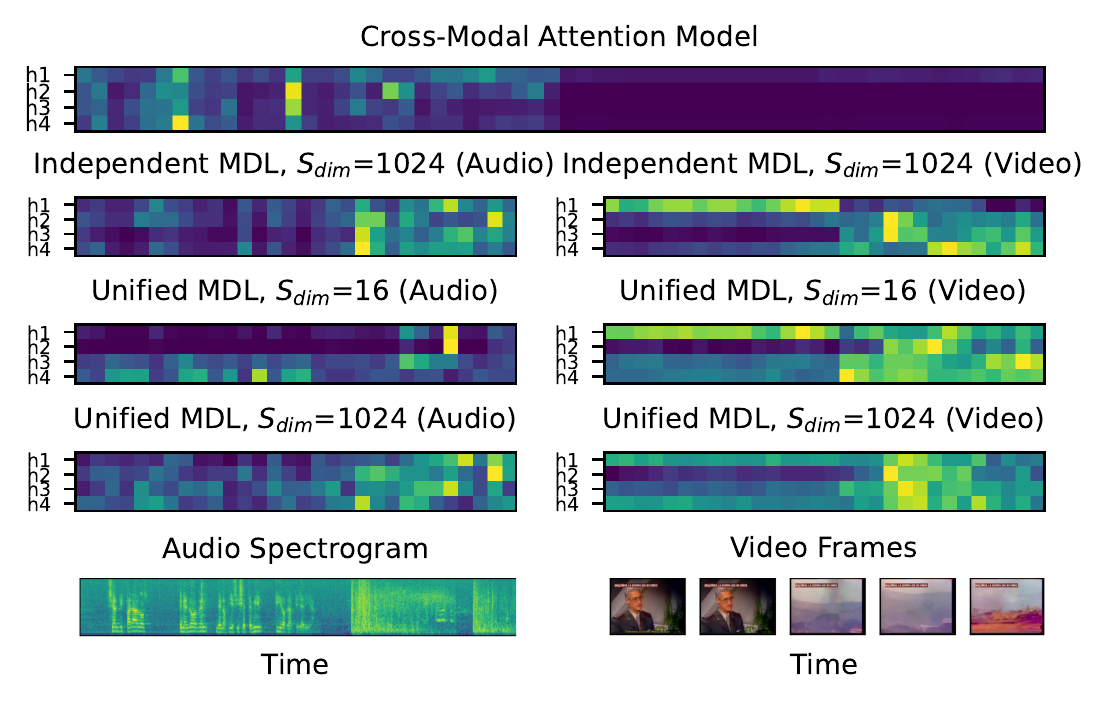}
\endminipage
\vspace{-0.5em}
\caption{(Left) Difference between the last layer attention maps with (paired) audio and video input for UAVM with various $N_s$. (Right) Temporal attention heatmap of 4 attention heads for a sample input (label: firing cannon). }
\label{fig:exp4}
\vspace{-1.0em}
\end{figure}

\squeezeup\squeezeup
\subsection{Attention Maps}

Even when the audio and video features are temporally synchronized, the informative part for event classification could be different. We show the temporal attention heatmap of 4 attention heads of the last Transformer layer in Figure~\ref{fig:exp4} (right). The UAVM, even when the shared Transformer has very limited capacity ($S_{dim}=16$), can pay attention to different parts of the audio and video, demonstrating its ability to process two modalities simultaneously. A cross-modal attention model, however, could be dominated by one modality while the other modality still contains useful information. Interestingly, by quantitatively calculating the mean absolute error (MAE) between attention maps of paired audio and video inputs, we find that the more modal-specific layers, the smaller the difference between the audio and video attention maps (Figure~\ref{fig:exp4}, left), indicating the modal-specific Transformers layers are forced to temporally align the informative part of the two modalities, so that succeeding shared Transformer layers can have a relatively more modal-agnostic attention map. 

\squeezeup
\section{Conclusion}

\rev{In this work, we unify the audio and visual branches of a multi-modality model and build UAVM. Performance-wise, UAVM is similar to a modal-independent model, and outperforms cross-modal attention models on VGGSound. We find the capacity of the weight-sharing network greatly impacts the behavior of UAVM. When its capacity is constrained, UAVM shows some intriguing unique properties, e.g., it maps audio and video in a unified latent space, and better aligns paired audio and video in the latent space. Such unique properties gradually disappear with increasing model capacity, and UAVM behaves closer to modal-independent models.}

\section{Acknowledgments and Disclosure of Funding}
\rev{This research is supported by the MIT-IBM Watson AI Lab.}

\bibliographystyle{IEEEtran}
\bibliography{ref}

\end{document}